\pdfoutput=1

\documentclass[11pt]{article}

\usepackage[final]{acl}

\usepackage{times}
\usepackage{latexsym}

\usepackage[T1]{fontenc}

\usepackage[utf8]{inputenc}

\usepackage{microtype}

\usepackage{inconsolata}

\usepackage{graphicx}

\usepackage{algorithm}
\usepackage{algpseudocode}
\usepackage{amsmath}
\usepackage{bbm}
\usepackage{blindtext}
\usepackage{enumitem}
\usepackage{makecell}
\usepackage{tcolorbox}
\usepackage{url}
\usepackage{footnote}

\usepackage{hyperref}

\DeclareMathOperator*{\argmin}{arg\,min}
\newcommand*\samethanks[1][\value{footnote}]{\footnotemark[#1]}
\newcounter{footnotesintable}
\newtcolorbox{mybox}{fontupper=\footnotesize}

\title{\centering SLIM: Subtrajectory-Level Elimination for More Effective Reasoning}

\author{
    \textbf{Xifeng Yao\thanks{~~Co-first authors}}, \textbf{Chengyuan Ma\samethanks}, \textbf{Dongyu Lang\samethanks}, 
    \textbf{Yinhao Ni}, \textbf{Zhiwei Xu}, \textbf{Huarui Xie},\\ \textbf{Zihao Chen}, 
    \textbf{Guang Shen}, \textbf{Dandan Tu}, \textbf{Yi Bai}, \textbf{Changzheng Zhang\thanks{~~Corresponding author}}\\
    Huawei Technologies Co., Ltd.\\
    \texttt{\{yaoxifeng, machengyuan1, zhangzhangzheng\}@huawei.com}\\
    \texttt{langdongyu@h-partners.com}
}

\begin{document}
\maketitle
\begin{abstract}
In recent months, substantial progress has been made in complex reasoning of Large Language Models (LLMs), particularly through the application of test-time scaling. Notable examples include, though are not limited to, OpenAI's o1/o3/o4 series and DeepSeek-R1. When responding to a query, these models generate an extended reasoning trajectory, during which the model explores, reflects, backtracks, and self-verifies before arriving at a conclusion. However, fine-tuning models with such reasoning trajectories may not always be optimal. Our findings indicate that not all components within these reasoning trajectories contribute positively to the reasoning process; in fact, some components may affect the overall performance negatively. In this study, we divide a reasoning trajectory into individual subtrajectories and develop a "5+2" framework to: (1) systematically identify suboptimal subtrajectories within the reasoning trajectory based on five human-established criteria; (2) assess the independence of the suboptimal subtrajectories identified in (1) from the subsequent content, ensuring that their elimination does not compromise overall flow and coherence of the reasoning process. Additionally, a sampling algorithm, built upon the "5+2" framework, is employed to select data whose reasoning process is free from suboptimal subtrajectories to the highest degree. Experimental results demonstrate that our method can reduce the number of suboptimal subtrajectories by 25.9\% during the inference. Furthermore, our method achieves an average accuracy of 58.92\% on highly challenging AIME24, AIME25, AMC24 and MATH500 benchmarks with only two thirds of training data, surpassing the average accuracy of 58.06\% achieved with the entire data, and outperforming open-source datasets, including s1K-1.1, Light-R1-SFT-stage-1, OpenR1-Math-94k, and OpenThoughts-114k, when fine-tuning Qwen2.5-Math-7B. Finally, we have validated the efficacy of our method under resource-constrained scenarios, where it exhibits performance improvements across different maximum inference token limits: 2k, 4k, 8k, and 16k tokens.

\end{abstract}

\section{Introduction}
Large language models (LLMs) have been rapidly evolving in their ability to tackle complex reasoning tasks. Recently, in the domain of LLMs, Reinforcement Learning (RL) employing an outcome-based reward has attracted public attention, as it grants the model extensive freedom to explore, reflect, backtrack, and self-verify, a process known as test-time scaling \citep{DeepSeekAI2025DeepSeekR1IR, deepscaler2025}. RL-ed LLMs, exemplified by DeepSeek-R1 \citep{DeepSeekAI2025DeepSeekR1IR}, have demonstrated robust capabilities in handling complex reasoning tasks, and are consequently often used as teacher models in knowledge distillation, enhancing the reasoning capabilities in other models or cold-starting them with a test-time scaled output format through Supervised Fine-Tuning (SFT) \citep{DeepSeekAI2025DeepSeekR1IR, Yang2025Qwen3TR, Wen2025LightR1CS}. However, reponses generated by RL-ed LLMs do not always guarantee the highest quality, as the unconstrained freedom during the RL training can introduce inefficiencies or counterproductive elements, such as prematurely abandoned steps or repetitive verifications, even within correct solutions, which will be illustrated in Section~\ref{section:A Deep Dive into Subtrajectories}. Fine-tuning a model using such solutions would be suboptimal, as they could potentially decrease both the model's accuracy \citep{Ye2025LIMOLI} and thinking efficacy \citep{Wang2025ThoughtsAA}.

Naturally, the following question arises: given a set of QA pairs, where the answers are distilled from a RL-ed LLM, how can we select the QA pairs that are free from these inefficiencies and counterproductive elements to fine-tune another model? To address this question, we first recall that answers from a RL-ed LLM, such as DeepSeek-R1, typically take the form in Appendix \ref{sec:Example of DeepSeek-R1-distilled QA Pairs}. 

For a QA pair, we divide its thinking process into individual approaches, referred to as \textbf{subtrajectories} in later discussions. We propose five criteria: \textit{Effort}, \textit{Effectiveness}, \textit{Coherence}, \textit{Preliminary Conclusion}, \textit{Valid Verification}, to assess each subtrajectory, determining whether it contributes positively to problem-solving from a specific perspective. If a subtrajectory fails to meet a criterion, we will further assess its independence within the thinking process and determine whether it can be removed without impacting the understanding and coherence of subsequent reasoning process. After eliminating suboptimal and independent subtrajectories, we will assign a quality score to the QA pair: first, assign a score to each subtrajectory based on the five criteria; second, aggregate these scores with weights proportional to the number of tokens in each subtrajectory. 

In addition to data quality, our analysis reveals that the distribution of the number of subtrajectories within the dataset also influences the model's reasoning ability. Accordingly, following the modification and computation of the quality scores, we develop a sampling algorithm of selecting QA pairs for supervised fine-tuning. This algorithm considers both the quality scores and the number of subtrajectories in the thinking process, achieving a balance through weights determined by the Kullback-Leibler (KL) divergence \citep{Joyce2011KullbackLeiblerD} between the distribution of number of subtrajectories in the entire dataset and that in the sampled dataset. This approach enables the selection of efficient and productive QA pairs based on an in-depth assessment of their thinking processes, while preventing the algorithm from disproportionately favoring thinking process with fewer subtrajectories.

Comprehensive experimental results illustrate that our methods achieve an average accuracy of 58.92\% on the highly challenging AIME24, AIME25, AMC24, and MATH benchmarks, utilizing merely two-thirds of the curated training data. This performance surpasses the 58.06\% accuracy obtained with the full dataset. Meanwhile, the number of suboptimal subtrajectories decreases by 25.9\% during the inference, which suggests a more profound and efficient reasoning paradigm. The concurrent enhancement in accuracy and thinking efficacy underscores that the responses generated from RL-ed LLMs indeed exhibit significant quality issues, and our data quality pipeline, encompassing suboptimal subtrajectory elimination and sampling strategy, demonstrates a robust capability in mitigating these issues.

In summary, our contributions are: (1) We propose a "5+2" framework to assess and modify the thinking processes generated from RL-ed LLMs at subtrajectory level. (2) We develop a sampling algorithm aimed at selecting efficient and productive QA pairs for supervised fine-tuning based on subtrajectory assessment. (3) We conduct comprehensive experiments and ablation studies to illustrate the effectiveness of `5+2'' framework and samping strategy, which enhance both model accuracy and thinking efficacy.

\section{Data Curation}

In this section, we discuss the process of constructing OpenSourceR1-Hard and DeepMath-Hard, the source datasets that we use for our subsequent studies. It should be noted that our hypotheses and methodologies in section \ref{sec: Sampling at the Subtrajectory Level} are both formulated and validated using OpenSourceR1-Hard, and the DeepMath-Hard dataset is regarded as an out-of-distribution test set. Both datasets undergo the following filtering processes, including basic quality filtering and difficulty filtering. The construction and filtration processes of the two datasets are elaborately detailed in Appendix \ref{sec:Data Curation}. We also decontaminate the collected dataset against the evaluation benchmarks mentioned in \ref{sec:setup} using 15-grams.

\section{Sampling at the Subtrajectory Level}
\label{sec: Sampling at the Subtrajectory Level}
\subsection{A Deep Dive into Subtrajectories}
\label{section:A Deep Dive into Subtrajectories}

When responding to a query, DeepSeek-R1, along with several RL-ed LLMs, initiates a thinking process. During this process, the model explores multiple approaches \citep{Qin2024O1RJ}, reflects on the current approach, reverts to previous steps when the current approach is longer deemed viable, and conducts self-verification. The attempted approaches, hereafter referred to as \textbf{subtrajectories}, are demarcated clearly, typically initiating with phrases such as "Alternatively", "Another method", and similar expressions. However, the quality of these subtrajectories is inconsistent, which in turn impacts the overall quality of the thinking process. After reviewing dozens of thinkings from OpenSourceR1-Hard, we identify that low-quality subtrajectories frequently manifest in the following forms (see Appendix \ref{sec:Examples of 5 types of subtrajetories} for examples):

\begin{enumerate}
    \item \textit{The subtrajectory proposes a method without attempting it.}
    \item \textit{The subtrajectory attempts to solve the problem in an ineffective manner.}
    \item \textit{The subtrajectory has logical discontinuities.}
    \item \textit{The subtrajectory transitions to the next one without reaching any conclusions.}
    \item \textit{The subtrajectory contains redundant self-verification(s).}
\end{enumerate}

We prompted QwQ-32B \citep{qwq32b} to assess whether subtrajectories in the OpenSourceR1-Hard dataset exhibit any of the aforementioned issues. The evaluation result revealed that 50.16\% of all subtrajectories contain at least one of the five defined low-quality characteristics.

\subsection{Identifying and Eliminating Suboptimal Subtrajectories}
\label{subsec: Identifying and Eliminating Suboptimal Subtrajectories}

To identify the five inefficient and counterproductive components, we have established five specific criteria. We prompt QwQ-32B with these criteria to evaluate each subtrajectory. These five criteria form the "+5" component of our "5+2" framework. 

\begin{enumerate}
    \item \textit{Effort}: The subtrajectory should not only introduce a method but also demonstrate its relevance to the current context. This involves providing a detailed explanation of the method and then applying it to address the problem at hand, integrating it with the preceding discussion or the problem statement. 
    \item \textit{Effectiveness}: The subtrajectory should attempt the problem in an effective manner. This may involve: simplifying the problem, refining previously suggested steps, advancing the problem-solving process, clarifying the limitations of the applied methods, or substantiating earlier conclusions. 
    \item \textit{Coherence}: Each step within the subtrajectory is logically connected, ensuring no logical leaps occur in the reasoning process. Every intermediate result must be derived through computation or rigorous proof. 
    \item \textit{Preliminary Conclusion}: Before transitioning to the next subtrajectory, this subtrajectory should draw a preliminary conclusion, which may include a final answer, intermediate findings, an evaluation of the current approach, or suggestions of other viable approachs. 
    \item \textit{Valid Verification}: The subtrajectory avoids repetitive verification of the same statement using the identical method, and it does not re-verify statements that have been verified in previous subtrajectories. 
\end{enumerate}

If a subtrajectory fails to meet any of the five criteria, it is classified as a \textbf{suboptimal subtrajectory}. The existence of suboptimal subtrajectories can degrade the overall quality of the thinking process. However, discarding QA pairs that include any suboptimal subtrajectory would significantly reduce the data size available for supervised fine-tuning. Instead, we opt to eliminate any identified suboptimal subtrajectory within the thinking process utilizing the five criteria. 

When eliminating subtrajectories, it is crucial to maintain the overall flow and structure of the thinking process. An example of non-eliminable suboptimal subtrajectory is shown in Appendix \ref{box:Example of Non-eliminable Suboptimal Subtrajectory}. Note that the first subtrajectory in the example fails to attempt the approach it proposes, thus violating the first criterion. However, this subtrajectory cannot be eliminated because the area fomula derived in it is revisited in the third subtrajectory, in which a valid attempt is made. Given this dependency, the first subtrajectory must be retained. 

Therefore, subtrajectories that are suboptimal should not be removed if their removal impairs the understanding of the subsequent content. To be more precise, upon identifying a suboptimal subtrajectory, we will prompt QwQ-32B to evaluate its independence from subsequent subtrajectories. Should this suboptimal subtrajectory be determined to be independent, it will be subject to elimination. 

\begin{enumerate}
    \item \textit{Independence}: Assessing whether the parameters, variables, algebraic expressions, conclusions, or verifications defined in the current subtrajectory are used in later content.
    \item \textit{Elimination}: If the current subtrajectory is relied upon by subsequent content, it should be retained. Conversely, if a subtrajectory is suboptimal and independent of subsequent subtrajectories, it should be eliminated. 
\end{enumerate}

This independence assessment and elimination mechanism constitutes the "+2" component of our "5+2" framework.

\subsection{The Sampling Algorithm}
\label{subsec:The Sampling Algorithm}
\subsubsection{Scoring a Thinking Process}
Due to the existence of suboptimal subtrajectories that cannot be eliminated, the \textbf{revised thinking process}, i.e., thinking process after elimination of independent suboptimal subtrajectories, cannot be problem-free. Therefore, we introduce a scoring mechanism to assess the quality of the revised thinking process, in accordance with the five criteria outlined in the preceding section. This scoring mechanism will be instrumental in the selection of QA pairs for supervised fine-tuning. 

Given a QA pair, we extract its thinking process. Next, we prompt QwQ-32B to evaluate each subtrajectory within this thinking process against the five criteria, with the aim of identifying and eliminating those suboptimal subtrajectories that are independent, as detailed in Section \ref{subsec: Identifying and Eliminating Suboptimal Subtrajectories}. Each of the remaining subtrajectories is awarded $\frac{1}{5}$ points for each of the five criteria it satisfies:

\begin{equation}
  \label{eq:score_subtrajectory}
  \scalebox{0.9}{$
  \begin{split}
        &Score\text{(subtrajectory)}:= \\
        &\sum_{j=1}^{5}\frac{1}{5} \cdot \mathbbm{1}\Bigl[\text{subtrajectory satisfies }\text{criterion}_j\Bigr].
  \end{split} $}
\end{equation}

Note that a score ranging from 0 to 1 is assigned to each subtrajectory. We will aggregate these individual scores into a single score that accurately reflects the overall quality of the thinking process.

\smallskip 

\subsubsection{Varied Weights Based on Token Counts}
\label{subsubsec:Varied Weights Based on Token Counts}
The length of subtrajectories is a critical factor. For longer suboptimal subtrajectories in the revised thinking process, a larger penalty should be imposed in contrast to their shorter counterparts. Consequently, when aggregating the scores of each subtrajectory, we apply a weight that is determined by the token count of the respective subtrajectory:

\begin{equation}
  \label{eq:quality_score2}
  \scalebox{0.9}{$
  \begin{split}
        &QualityScore\text{(thinking)} := \\
        &\sum_{i=1}^{n} \frac{T(\text{subtrajectory}_i)}{T(\text{thinking})} \left(Score\text{(subtrajectory)}\right),
  \end{split}$}
\end{equation}

where $n$ is the number of subtrajectories within the thinking process, and $T(\cdot)$ returns the number of tokens of the input string. The flow chart in Appendix \ref{sec:Demonstration of Varied Weights Based on Token Counts} demonstrates the computation process of the varied weights based on token counts.

Technically speaking, the quality score is specifically defined on the thinking process within the answer of a QA pair. Given that each QA pair contains exactly one thinking process, we will adopt a less rigorous notation: \textit{QualityScore}(QA pair), to denote the quality score of the thinking process within the answer of that QA pair.

\subsubsection{Sampling on Quality Score and Distribution of Subtrajectory Counts}
Naturally, after calculating the quality score of a thinking process, we can establish a threshold and select QA pairs whose thinking process scored above this threshold. However, we notice that the scoring mechanism disproportionately favors thinking processes with fewer subtrajectories, as they are less prone to violate criterion 1, 2, 4. The findings are detailed in Appendix \ref{sec: Distribution of Number of Subtrajectories after Sampling on Quality Score}.

Theoretically speaking, a SFT dataset comprising an excessive number of QA pairs with an extremely low number of subtrajectories may lead to a reduction in the SFTed model's exploratory ability, confining its search to a limited space and thereby impairing its performance on complex reasoning tasks. Therefore, when sampling based on quality scores, it is essential to introduce a constraint by incorporating a penalty term that reflects the percentage change in the frequency of number of subtrajectories within the thinking process of the sampled dataset and the entire dataset. The detailed sampling algorithm is in Appendix \ref{sec: sampling algorithm}.

Through the sampling algorithm, we may select QA pairs that are aligned with the five criteria outlined in Section \ref{subsec: Identifying and Eliminating Suboptimal Subtrajectories}, while considering the number of subtrajectories as intact as possible. 

\section{Experimental Results}

\subsection{Setup}

\label{sec:setup}
\textbf{Training}: We conduct supervised fine-tuning on Qwen2.5-Math-7B across two datasets: OpenSourceR1-Hard and DeepMath-Hard to evaluate the effectiveness of our methods in the domain of mathematics. The detailed training configurations is in Appendix \ref{sec:Training configurations}.

\noindent \textbf{Evaluation}: We assess the effectiveness of our methods using a range of mathematics benchmarks, including AIME24, AIME25, MATH500, AMC24, as detailed in Appendix \ref{sec:Benchmarks}. The evaluation methods are detailed in Appendix \ref{sec:Evaluation Methods}.

\subsection{Ablation Studies}

We conduct ablation studies to assess the efficacy of the "5+2" framework and the sampling algorithm on both our in-distribution dataset OpenSourceR1-Hard, and our out-of-distribution dataset DeepMath-Hard. Regarding the OpenSourceR1-Hard dataset (around 60k samples), we have curated various fractions of the dataset, including the entire dataset, two-thirds of the dataset, and one-third of the dataset. Regarding the DeepMath-Hard dataset (around 12k samples), we have curated two fractions: the entire dataset and two-thirds of the dataset. Due to the relatively limited size of the DeepMath-Hard dataset, we did not curate a one-third fraction in our analysis. 

For each sampled fraction, we consider the following four configurations, as detailed in Appendix \ref{sec:four configurations}: (1) Elimination with Sampling Algorithm (\textit{E+SA}); (2) No Elimination with Sampling Algorithm (\textit{NE+SA}); (3) Elimination without Sampling Algorithm (\textit{E+NSA}); (4) No Elimination without Sampling Algorithm (\textit{NE+NSA}).

For the entire dataset, only the configurations \textit{E+NSA} and \textit{NE+NSA} are employed, as the sampling algorithm is inapplicable in this context.

The performance of the OpenSourceR1-Hard models is detailed in Table \ref{tab:OpenSourceR1-Hard Solution Quality}. The results demonstrate that the elimination of suboptimal subtrajectories enhances the model's performance across all comparative groups, regardless of the application of the sampling algorithm. Specifically, within the entire dataset, the \textit{E+NSA} configuration achieves an accuracy of 59.60\%, outperforming the \textit{NE+NSA} configuration, which attains 58.06\%. Similarly, in the two-thirds of the dataset, when the sampling algorithm is applied, \textit{E+SA} achieves an accuracy of 58.92\%, representing a 1.86\% improvement over \textit{NE+SA}. This enhancement can be attributed to the efficacy in eliminating suboptimal subtrajectories, thereby optimizing the overall solution's efficiency despite a reduction in token length. In the one-third of the dataset, elimination suboptimal subtrajectories achieves approximately the same accuracy as configurations without the elimination process. To our best knowledge, this similarity in performance is partly due to the 7B model's limited math capabilities compared to larger models such as the 32B variant, making its performance highly susceptible to the quantity of data and tokens utilized in the SFT process. Despite eliminating suboptimal subtrajectories further reducing the number of tokens used in SFT, it manages to maintain a comparable level of accuracy to configurations with the original solution.

Moreover, the integration of the "5+2" framework with sampling algorithms demonstrates a pronounced capability in augmenting model performance. Specifically, the implementation of \textit{E+SA} significantly enhances model accuracy from 56.23\% (as observed in \textit{NE+NSA}) to 58.92\% in the two-third of the dataset. A similar observation has been made within the one-third of the dataset. Additionally, the \textit{E+SA} model in the two-third of the dataset demonstrates a 0.86\% better performance compared to the \textit{NE+NSA} model in the entire dataset. This suggests that although reductions in sample size and token amounts can significantly influence a 7B model in SFT process, the "5+2" framework together with the sampling algorithm are particularly effective in identifying optimal QA pairs from the entire dataset, thereby achieving enhanced performance.

\begin{table}[h]  
    \centering
    \resizebox{\columnwidth}{!}{
    \begin{tabular}{cccccc}  
        \Xhline{2\arrayrulewidth}
        \textbf{Methods} & \textbf{AIME25} & \textbf{AIME24} & \textbf{MATH500} & \textbf{AMC24} & \textbf{Average}\\
        \hline
        \multicolumn{6}{c} {\centering Entire Dataset}\\
        \hline  
        E+NSA & 35.03 & 44.15 & 90.25 & 68.98 & \textbf{59.60}\\
        NE+NSA & 29.18 & 47.50 & 88.90 & 66.65 & 58.06\\
        \hline
        \multicolumn{6}{c} {\centering Two-thirds of the Dataset}\\
        \hline 
        E+SA & 38.63 & 39.43 & 90.55 & 67.05 & \textbf{58.92}\\
        NE+SA & 35.85 & 36.70 & 89.80 & 65.90 & 57.06\\
        E+NSA & 37.50 & 38.35 & 89.40 & 60.40 & 56.41\\
        NE+NSA & 31.65 & 35.80 & 89.25 & 68.23 & 56.23\\
        \hline
        \multicolumn{6}{c} {\centering One-third of the Dataset}\\
        \hline 
        E+SA & 29.45 & 35.00 & 87.20 & 66.30 & \textbf{54.49}\\
        NE+SA & 30.55 & 35.00 & 87.05 & 65.33 & 54.48\\
        E+NSA & 27.50 & 34.15 & 87.95 & 60.98 & 52.65\\
        NE+NSA & 27.50 & 33.90 & 87.90 & 61.73 & 52.76\\
        \Xhline{2\arrayrulewidth}
    \end{tabular}
    }
    \caption{OpenSourceR1-Hard: The "5+2" framework and the sampling algorithm performance across mathematical benchmarks} 
    \label{tab:OpenSourceR1-Hard Solution Quality}
\end{table}

A similar trend is observed on our out-of-distribution dataset DeepMath-Hard, as summarized in Table \ref{tab:DeepMath Solution Quality}. Specifically, within the entire subset, the \textit{E+NSA} configuration achieved an accuracy of 52.53\%, significantly surpassing the 50.21\% accuracy of the \textit{NE+NSA} configuration. Moreover, in the two-third of the dataset, the implementation of \textit{E+SA} yielded an accuracy rate of 49.12\%, outperforming the 47.05\% achieved by \textit{NE+NSA}. These findings indicate that, even when evaluated on an out-of-distribution dataset, the integration of the "5+2" framework and the sampling algorithm exhibits superior performance across various data sizes, outperforming configurations that do not incorporate these methods.

\begin{table}[h]  
    \centering
    \resizebox{\columnwidth}{!}{
    \begin{tabular}{cccccc}  
        \Xhline{2\arrayrulewidth}
        \textbf{Methods} & \textbf{AIME25} & \textbf{AIME24} & \textbf{MATH500} & \textbf{AMC24} & \textbf{Average}\\
        \hline
        \multicolumn{6}{c} {\centering Entire Dataset}\\
        \hline  
        E+NSA & 29.45 & 28.60 & 87.65 & 64.40 & \textbf{52.53}\\
        NE+NSA & 25.00 & 28.60 & 87.40 & 59.85 & 50.21\\
        \hline
        \multicolumn{6}{c} {\centering Two-thirds of the Dataset}\\
        \hline 
        E+SA & 27.23 & 27.23 & 85.20 & 56.80 & \textbf{49.12}\\
        NE+SA & 25.03 & 25.55 & 86.15 & 56.25 & 48.25\\
        E+NSA & 24.18 & 27.50 & 85.65 & 56.80 & 48.53\\
        NE+NSA &  20.55 & 26.38 & 85.05 & 56.23 & 47.05\\
        \Xhline{2\arrayrulewidth}
    \end{tabular}
    }
    \caption{DeepMath-Hard: The "5+2" framework and the sampling algorithm performance across mathematical benchmarks}  
    
    \label{tab:DeepMath Solution Quality}
\end{table}

In addition, we conducted two extra sets of ablation studies to validate our methods in Section \ref{sec: Sampling at the Subtrajectory Level}. The first ablation study compares equal weights, detailed in Appendix \ref{section: Equal Weights}, against varied weights. The second one contrasts the presence and absence of the sampling algorithm. The results of these two ablation sturdies are detailed in Appendix \ref{section: Ablation Studies}.

\subsection{Main Results}
\subsubsection{Comparison with Other Datasets}
In the ablation studies, we have curated datasets of varying sizes from OpenSourceR1-Hard, employing both our "5+2" framework and the sampling algorithm with target data size $d$ set to 1k, 20k, 40k, 60k, respectively. These datasets are labeled as \textit{OpenSourceR1-Hard E+SA (1k)}, \textit{OpenSourceR1-Hard E+SA (1/3)}, \textit{OpenSourceR1-Hard E+SA (2/3)}, \textit{OpenSourceR1-Hard E+SA (1)}, in ascending order of their size.

We benchmark our four sampled datasets derived from OpenSourceR1-Hard against several established open-source datasets (as shown in Appendix \ref{sec:sampled_dataset}). This evaluation was performed by fine-tuning the Qwen2.5-Math-7B model under the training configurations detailed in subsection \ref{sec:setup}.

\begin{table}[h]  
    \centering
    \resizebox{\columnwidth}{!}{
    \begin{tabular}{ccccccc}  
        \Xhline{2\arrayrulewidth}
        \textbf{Dataset} & \textbf{Size} & \textbf{AIME25} & \textbf{AIME24} & \textbf{MATH500} & \textbf{AMC24} & \textbf{Average}\\
        \hline  
        s1K-1.1 & 1k & 10.83 & 18.08 & 77.15 & 37.33 & 35.85\\
        \textit{OS-R1-H E+SA (1k)} & 1k & 16.65 & 18.35 & 75.15 & 39.80 & \textbf{37.49}\\
        \hline
        Light-R1-SFT-stage-1 & 76k & 33.05 & 39.45 & 88.65 & 65.53 & 56.67\\
        OpenR1-Math-94k & 94k & 30.55 & \textbf{46.10} & 88.95 & 64.58 & 57.55\\
        OpenThoughts-114k & 114k & 29.45 & 35.28 & 88.85 & 62.88 & 54.12\\
        \textit{OS-R1-H E+SA (1/3)} & 20k & 29.45 & 35.00 & 87.20 & 66.30 & 54.49\\
        \textit{OS-R1-H E+SA (2/3)} & 40k & \textbf{38.63} & 39.43 & \textbf{90.55} & 67.05 & 58.92\\
        \textit{OS-R1-H E+SA (1)} & 60k & 35.03 & 44.15 & 90.25 & \textbf{68.98} & \textbf{59.60}\\
        \Xhline{2\arrayrulewidth}
    \end{tabular}
    }
    \caption{Comparison of Our Datasets with Other Datasets. OS-R1-H stands for OpenSourceR1-Hard} 
    \label{tab:comparison}
\end{table}

Note that our sampled datasets achieve superior performance compared to all selected open-source datasets, despite being only a fraction of their size. Furthermore, our methods demonstrate efficacy even when applied to considerably smaller datasets. Specifically, the dataset comprising 1,000 instances achieved an accuracy rate of 37.49\%, outperforming the 35.85\% from s1k-1.1, which is a meticulously curated collection of 1,000 instances through rigorous refinement processes.

\subsubsection{Analysis of underthinking phenomenon}
In addition to evaluating model accuracy, recent studies have identified an "underthining" phenomenon \citep{Wang2025ThoughtsAA} in o1-like LLMs, where the model frequently switches between reasoning trajectories without sufficiently exploring each one. Our "5+2" framework, coupled with the sampling algorithm, is specifically designed to eliminate suboptimal subtrajectories and filter out QA pairs that contain such suboptimal subtrajectories. Therefore, we hypothesize an improvement in model's ability to respond to questions with a reduced number of subtrajectories and a deeper analysis within each subtrajectory. To validate our hypothesis, we analyze the variations in total number of tokens in the reasoning process, the number of subtrajectories and the average number of tokens per subtrajectory before and after fine-tuning with the following datasets:
\begin{itemize}
    \item \textit{OpenSourceR1-Hard E+SA (2/3)} and \textit{DeepMath-Hard E+SA (2/3)}, datasets where both the "5+2" framework and the sampling algorithm are applied.
    \item \textit{OpenSourceR1-Hard NE+NSA (2/3)} and \textit{DeepMath-Hard NE+NSA (2/3)}, randomly selected datasets with no additional operation.
\end{itemize}

\begin{figure*}[h]
    \centering
    \includegraphics[scale=0.5]{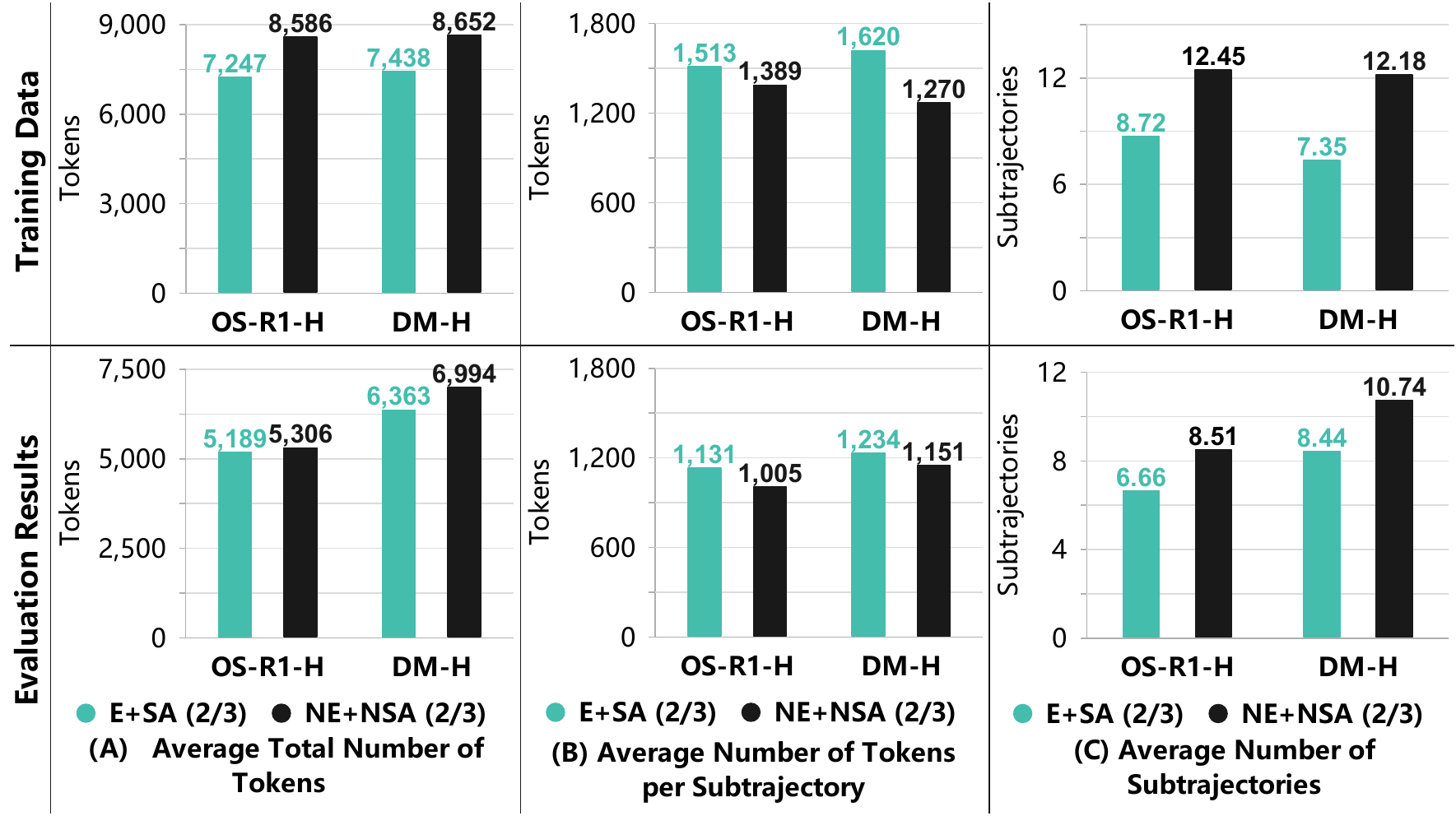}
    \caption{Comparison of Metrics for Thinking Efficacy between Training Data and Evaluation Results, where OS-R1-H stands for OpenSourceR1-Hard and DM-H stands for DeepMath-Hard.}
    \label{all_diagrams}
\end{figure*}

In Figure \ref{all_diagrams} (A), we observe a notable reduction in the total number of tokens involved in the reasoning process within the training datasets when the "5+2" framework and sampling algorithm are employed. For the OpenSourceR1-Hard dataset, the total number of tokens decreased by 15.6\% (from 8,586 to 7,247), and for the DeepMath-Hard dataset, a 14.0\% reduction (from 8,652 to 7,438) was observed. When evaluating models fine-tuned on these datasets, we noticed a 2.2\% decrease (from 5,306 to 5,189) for OpenSourceR1-Hard and a more substantial 9.0\% decrease (from 6,994 to 6,363) for DeepMath-Hard, respectively.

In Figure \ref{all_diagrams} (B), in the training data, we observe an 8.9\% increase in the average number of tokens per subtrajectory, rising from 1,389 to 1,513 for OpenSourceR1-Hard when applying \textit{E+SA}. Similarly, DeepMath-Hard shows an 27.6\% increase under the same conditions. This phenomenon is also observed in the evaluation results post fine-tuning. Models fine-tuned with OpenSourceR1-Hard exhibits an average increase of 12.5\% in the average number of tokens per subtrajectory, rising from 1,005 to 1,131 tokens. Similarly, models fine-tuned with DeepMath-Hard shows a 7.2\% increase. This implies a deep thinking paradigm during the inference process.

In Figure \ref{all_diagrams} (C), the application of \textit{E+SA} is notably associated with a significant decrease in the average number of subtrajectories. On the OpenSourceR1-Hard dataset, the average number of subtrajectories decreases from 12.45 to 8.72, a 29.96\% reduction. A more pronounced decline: 39.66\% is observed on the DeepMath-Hard dataset, with the average dropping from 12.18 to 7.35. A consistent trend is also evident in the evaluation results. When \textit{E+SA} is both applied, a significant reduction in the average number of subtrajectories is observed for models that have been fine-tuned on the OpenSourceR1-Hard and DeepMath-Hard datasets. The average number of subtrajectories decreases by 21.74\%, from 8.51 to 6.66, for the OpenSourceR1-Hard dataset. Similarly, for the DeepMath-Hard dataset, the average number of subtrajectories is reduced by 21.41\%, decreasing from 10.74 to 8.44.

The empirically findings indicate that the "5+2" framework and the sampling algorithm, or equivalently, models fine-tuned with our datasets effectively mitigate the "underthinking" phenomenon. This is exemplified by a reduction in the number of subtrajectories, coupled with an increase in the number of tokens within each subtrajectory. This outcome signifies a decrease in the frequency of switching approaches and a deeper reasoning within each approach.

\subsubsection{Analysis of Suboptimal Subtrajectories in the Evaluation Results}

\begin{figure}[h]
    \centering
    \includegraphics[width=\columnwidth]{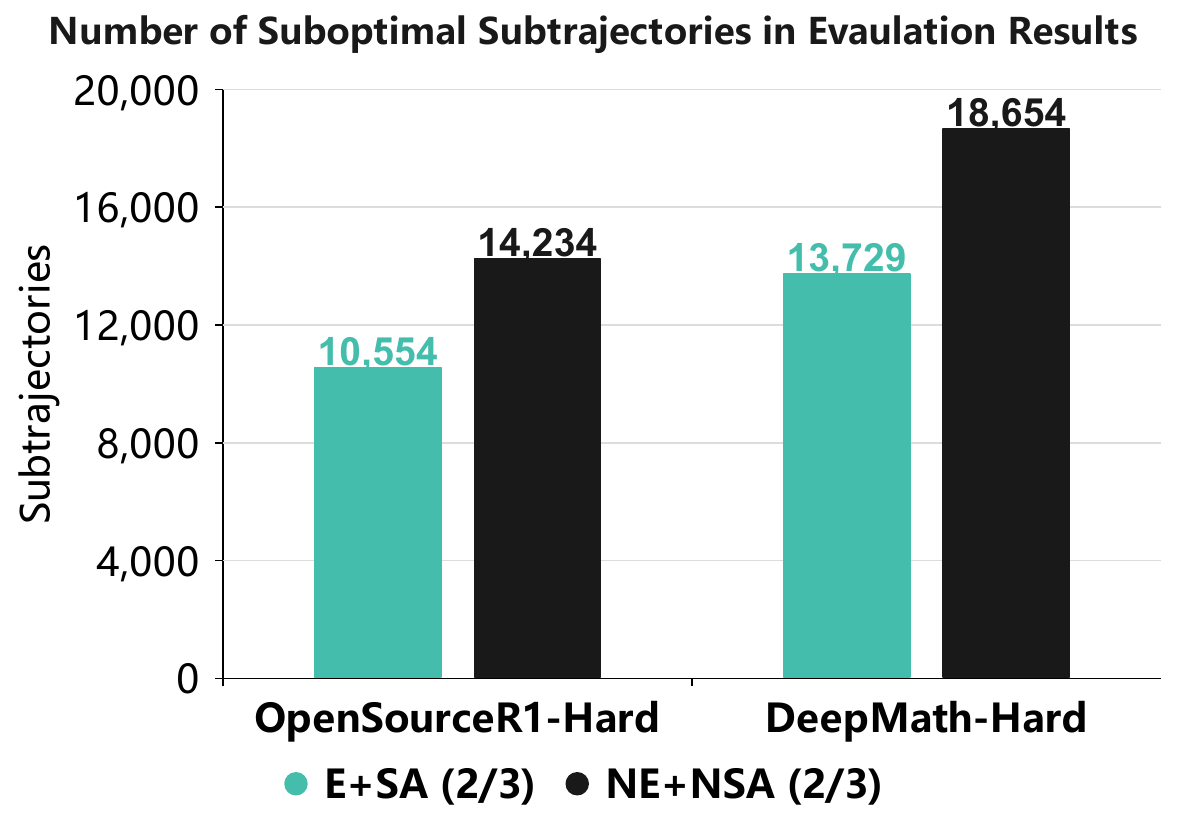}
    \caption{Average Number of Suboptimal Subtrajectories}
    \label{Average Number of Suboptimal Subtrajectories}
\end{figure}

One major aspect of our methods is that the model after fine-tunning with the "5+2" framework and the sampling algorithm is able to generate less number of suboptimal subtrajectories in the evaluation results. Specifically, in Figure \ref{Average Number of Suboptimal Subtrajectories}, model fine-tuned with OpenSourceR1-Hard has a 25.9\% (14,234 to 10,554) drop of the number of suboptimal subtrajectories, and with DeepMath-Hard, a 26.4\% (18,654 to 13,729) drop of the number of suboptimal subtrajectories with our method applied. See Appendix \ref{sec:Examples of Inference Outputs} for an example.

\subsubsection{The Effectiveness of Thinking Budget}
\begin{figure}[h]
    \centering
    \includegraphics[width=\columnwidth]{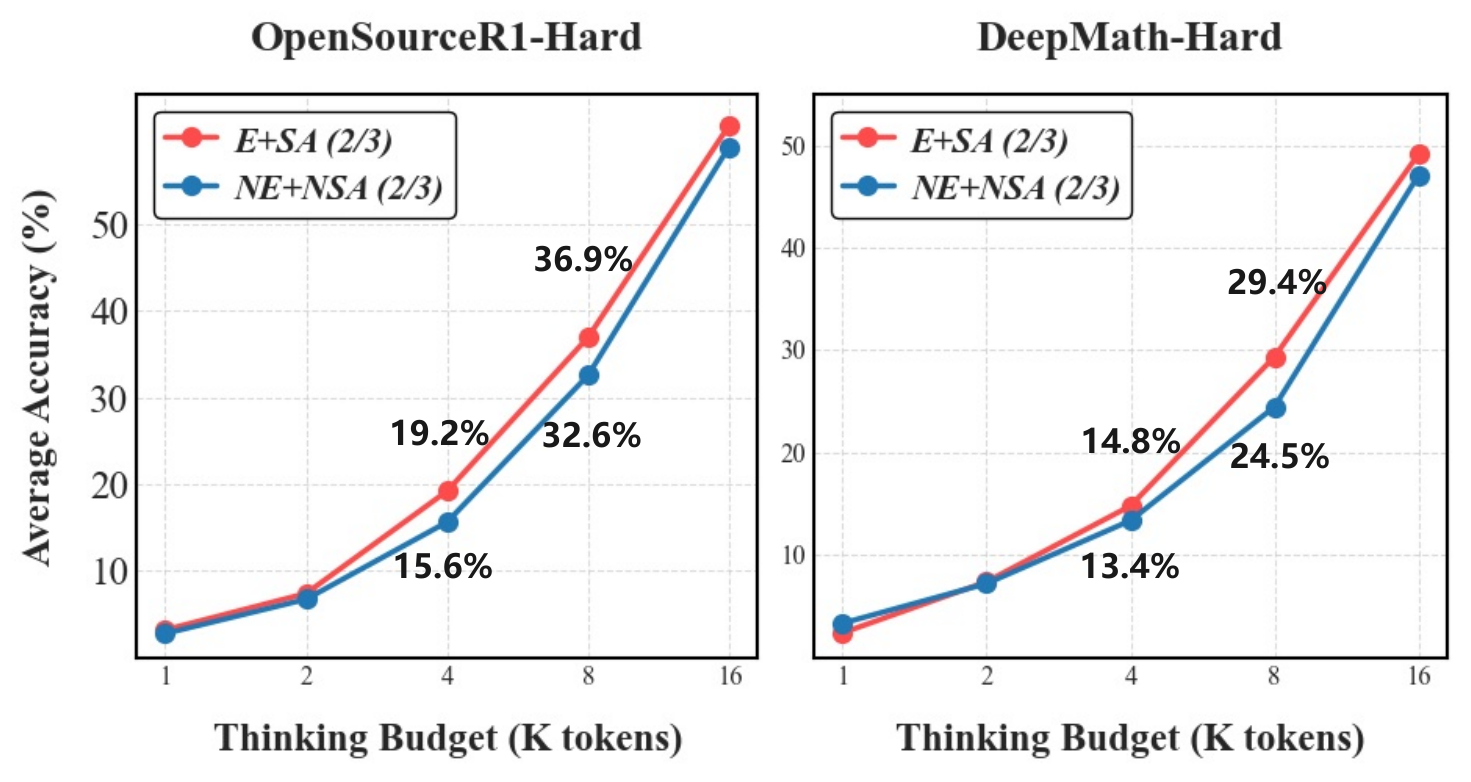}
    \caption{Accuracy of \textit{E+SA} and \textit{NE+NSA} with respect to the thinking budget.}
    \label{Accuracy of "E+SA" and "NE+NSA" with respect to the thinking budget.}
\end{figure}

To verify the effcetiveness of our method at different thinking budgets, we allocated 1k-16k thinking budgets on the four evaluation benchmarks. The resulting scaling curves are given in Figure \ref{Accuracy of "E+SA" and "NE+NSA" with respect to the thinking budget.}, \textit{E+SA} demonstrates a significant improvement over \textit{NE+NSA} across the 2k-16k budget range on both OpenSourceR1-Hard and DeepMath-Hard datasets. 

\section{Related Work}
\subsection{Test-Time Scaling}
Test-time scaling refers to the practice of enabling LLMs to generate a larger number of tokens during the inference phase, thereby significantly enhancing their problem-solving capabilities. Recent research in this area has primarily focused on two strategies\citep{OpenAI, Snell2024ScalingLT}: (1) Deploy LLMs to generate multiple reasoning trajectories, from which the optimal path is selected through the application of reward models\citep{Snell2024ScalingLT, Wu2024InferenceSL, Brown2024LargeLM}.  Such test-time scaling methods include parallel sampling\citep{Brown2024LargeLM, Wang2022SelfConsistencyIC} in which the majority voting mechanism is utilized to select the final answer from multiple generated solutions, and the tree-based search methods\citep{Yao2023TreeOT, Zhang2024AccessingGL, Qi2024MutualRM} like Monte-Carlo Tree Search (MCTS). (2) Employ reinforcement learning in the post-training of large LLMs, exemplified by models such as DeepSeek-R1, Qwen-QwQ\citep{Qwq}, DeepSeek-R1\citep{DeepSeekAI2025DeepSeekR1IR}, and Kimi-1.5\citep{Team2025KimiKS}. These models are capable of exploration, reflection, backtracking, and self-verification, therefore generating significantly longer outputs during inference time.

\subsection{Data Selection Policy}
It has been empirically demonstrated that high-quality data can enable LLMs to achieve optimal performance with a relatively small number of training samples\citep{Yang2024Qwen25TR, DeepSeekAI2024DeepSeekV3TR, Yu2023MetaMathBY}. For instance, s1\citep{Muennighoff2025s1ST} demonstrates that a 32B model trained on a dataset of 1,000 samples outperforms OpenAI's o1-preview. Similarly, LIMO\citep{Ye2025LIMOLI} substantiates the importance of data quality by employing three quality metrics: Optimal Structural Organization, Effective Cognitive Scaffolding, and Rigorous Verification in the selection of training data; a 32B model trained on a dataset of 819 samples, selected through these three criteria, surpasses the performance of o1-preview. 

\subsection{Thinking Efficacy} 
Reinforcement learning (RL) enhances model's ability to handle complex reasoning tasks by extending its reasoning process. However, during the problem-solving process, RL forms a unified and specific reasoning paradigm, regardless of the problem's complexity. This paradigm can become highly inefficient if not properly constrained\citep{Ye2025LIMOLI}. In the case of simpler problems, this paradigm may lead to overthinking, as these problems could be resolved with significantly less computational resources\citep{Chen2024DoNT}. Conversely, for more complex problems, this paradigm may introduce a significant number of ineffective and counterproductive elements into the reasoning process. Such elements not only compromise the model's accuracy but also diminish its token efficiency. One of such elements is underthinking, where the model switches between strategies too frequently without adequately exploring each one. To mitigate underthinking, \citep{Wang2025ThoughtsAA} proposes a decoding strategy that encourages a deeper exploration of each attempted strategy, thereby improving overall accuracy and thinking efficiency. In addition, \citep{Qiao2025ConCISECC} proposed a ConCISE framework to decrease redundant reasoning steps via confidence strategy in during inference time.

\section{Conclusion and Future Works}
In this paper, we conducted a comprehensive analysis of the quality of subtrajectories within the reasoning process of RL-LLMs. This analysis led to the identification of five critical quality issues that negatively impact both the accuracy and thinking efficacy of these models. To address these issues, we introduce a "5+2" framework to: (1) systematically identify suboptimal subtrajectories within the reasoning trajectory based on five human-established criteria; (2) assess the independence of the suboptimal subtrajectories identified in (1) from the subsequent content, ensuring that their elimination does not compromise overall flow and coherence of the reasoning process. Furthermore, we propose a sampling algorithm, built upon the "5+2" framework, to select data that are free from the identified quality issues to the maximum extent. Our experimental findings illustrate that our methods not only improve model accuracy but also enhances thinking efficacy by mitigating the "underthinking" issue, reducing the number of suboptimal subtrajectories, thereby improving the efficacy across different thinking budgets. In the future, we will generalize our method to other disciplines, such as physics and coding. Additionally, we aim to investigate the scalability of our framework by applying it to models of varying sizes, with larger models such as those with 32B parameters or more.

\section*{Limitations}

\begin{enumerate}
    \item It is worth noting that while our methods demonstrate significant utility in math domain, where multiple subtrajectories are often presented, domains with fewer number of subtrajectories or those that differ significantly in their reasoning paradigm may necessitate customized frameworks to attain comparable benefits.
    \item Our methods primarily emphasize data quality. Beyond quality, the diversity of the QA pairs is also a crucial factor. It is noteworthy that our methods designed to enhance quality may inadvertently lead to imbalances in the diversity distribution of the dataset.
\end{enumerate}

\bibliography{anthology, custom}

\clearpage

\appendix

\section{Example of DeepSeek-R1-distilled QA Pairs}
\label{sec:Example of DeepSeek-R1-distilled QA Pairs}
\begin{tcolorbox}[title={Example of DeepSeek-R1-distilled QA Pairs}]
    \textbf{Question:} Let the circles $k_{1}$ and $k_{2}$ intersect at two distinct points $A$ and $B$, and let $t$ be a common tangent of $k_{1}$ and $k_{2}$, that touches $k_{1}$ and $k_{2}$ at $M$ and $N$, respectively. If $t \perp A M$ and $M N=2 A M$, evaluate $\angle N M B$.

    \textbf{Think:} 
    
    <think>

    \textit{Alright}, let's try to tackle this problem step by step \dots

    \textit{Alternatively}, since $O_1M$ and $O_2N$ are both perpendicular to $t$ \dots

    \textit{Alternatively}, maybe the problem is from a math competition like the AIME, where $n$ is given, \dots
    
    \textit{Alternatively}, since $AM$ is perpendicular to $t$, and $MN$ is along $t$, then \dots

    \dots
    
    </think>

    \textbf{Final Answer:} Given two intersecting circles $k_1$ and $k_2$ with common points $A$ and $B$, ... The measure of $\angle NMB$ is $\backslash boxed\{45^\circ\}$.
\end{tcolorbox}

\section{Data Curation}
\label{sec:Data Curation}
\subsection{Filtering Processes}
\subsubsection{Basic Quality Filtering}
We design a rule-based filtering task, aiming to remove any entries that meet certain predefined criteria, including: (1) image-dependent questions: questions that require images, in forms of hyperlinks or visual references, to be answered; (2) truncated solutions: solutions that are cut off due to the length of the model output exceeding the predefined maximum token limit; (3) inconsistent language use: entries with mixed or incoherent language, such as abrupt shifts between English and Chinese.

\subsubsection{Difficulty Filtering}
We implement a two-stage difficulty filtering process, similar to the ones employed in s1 \citep{Muennighoff2025s1ST} and LIMO \citep{Ye2025LIMOLI}. The primary objective of this process is to only retain those entries that contain questions requiring highly complex and intricate reasoning solutions, thereby exceeding the capabilities of the current base models. For each entry (question, R1 solution), we deploy two models: Qwen2.5-Math-7B-Instruct \citep{Yang2024Qwen25MathTR} and R1-Distill-Qwen-7B \citep{DeepSeekAI2025DeepSeekR1IR}, to independently generate answers twice. Following the generation, we deploy a third 7B model, specifically fine-tuned for the purpose of assessing the correctness of generated answers in comparison to the ground truth. In this scenario, the 7B model evaluates the generated answers against the final answer extracted from the R1 solution. We exclude any entries where either model provides a correct answer at least once, and any entries in which there is no clearly marked final answer, i.e., boxed\{\}, in the R1 solution.

\subsection{Curation of OpenSourceR1-Hard}
We collect 5 open-source R1-distilled datasets from Hugging Face, totaling 210k samples after deduplication. The basic information of the collected datasets is listed as follows:

\begin{table}[h]
  \centering
  \begin{tabular}{lc}
    \hline
    \textbf{Dataset} & \textbf{Dataset Size} \\
    \hline
    OpenThoughts-114k\addtocounter{footnote}{1}\addtocounter{footnotesintable}{1}\footnotemark[\thefootnote] & 114k \\
    OpenR1-Math-94k\addtocounter{footnote}{1}\addtocounter{footnotesintable}{1}\footnotemark[\thefootnote] & 94k \\
    s1K-1.1\addtocounter{footnote}{1}\addtocounter{footnotesintable}{1}\footnotemark[\thefootnote] & 1k \\
    Light-R1-SFT-stage-1\addtocounter{footnote}{1}\addtocounter{footnotesintable}{1}\footnotemark[\thefootnote] & 76k \\
    Light-R1-SFT-stage-2\addtocounter{footnote}{1}\addtocounter{footnotesintable}{1}\footnotemark[\thefootnote] & 3k \\
    \hline
  \end{tabular}
  \caption{\label{citation-guide}
    Basic Information of Collected Datasets
  }
\end{table}

\addtocounter{footnote}{-\thefootnotesintable}
\addtocounter{footnote}{1}\footnotetext[\thefootnote]{\url{https://huggingface.co/datasets/open-thoughts/OpenThoughts-114k}}
\addtocounter{footnote}{1}\footnotetext[\thefootnote]{\url{https://huggingface.co/datasets/llamafactory/OpenR1-Math-94k}}
\addtocounter{footnote}{1}\footnotetext[\thefootnote]{\url{https://huggingface.co/datasets/simplescaling/s1K-1.1}}
\addtocounter{footnote}{1}\footnotetext[\thefootnote]{\url{https://huggingface.co/datasets/qihoo360/Light-R1-SFTData}}
\addtocounter{footnote}{1}\footnotetext[\thefootnote]{\url{https://huggingface.co/datasets/qihoo360/Light-R1-SFTData}}

After applying both basic quality filtering and difficulty filtering, we curated a dataset of 59,759 entries, which we will refer to as OpenSourceR1-Hard in later discussions. We remark that our hypotheses and methodologies presented in later sections are both formulated and validated using OpenSourceR1-Hard, thereby making it an in-distribution dataset.

\subsection{Curation of DeepMath-Hard}
During the preparation of this paper, we came across a recently released dataset called DeepMath \citep{He2025DeepMath103KAL}, a 103K R1-distilled dataset. We apply the same filtering processes to DeepMath, including both the basic quality filtering and difficulty filtering, albeit with slightly adjusted sampling parameters. This reduces the original DeepMath dataset to a more compact version, referred to as DeepMath-Hard, which consists of only 12,719 entries. The rationale for treating OpenSourceR1-Hard and DeepMath-Hard as separate datasets, rather than concatenating them, stems from the fact that our hypotheses and methodologies are both formulated and validated using OpenSourceR1-Hard. To evaluate the generalization of our approach, we deliberately isolated the DeepMath-Hard dataset, which will function as an out-of-distribution test set.

\section{Examples of 5 types of subtrajetories}
\label{sec:Examples of 5 types of subtrajetories}

\begin{enumerate}
    \item \textit{The subtrajectory proposes a method without attempting it.}
    \begin{tcolorbox}[title = {Example 1}, label=box:example1]
        \textit{Alternatively}, this is similar to a three-dimensional matching problem, which is NP-hard, but maybe in this specific case, with the constraints on the digit sums, it can be solved more easily.
    \end{tcolorbox}
    Example 1 mentions a three-dimensional matching problem; however, it falls short by not stating a precise definition of the problem, its relevance to the current context, and any attempt to address and resolve the problem through this approach.

    \item \textit{The subtrajectory attempts to solve the problem in an ineffective manner.}
    \begin{tcolorbox}[title = {Example 2}, label=box:example2]
        \textit{Alternatively}, \dots  Let's start testing\\ numbers step by step, starting from the\\ smallest natural numbers, checking if \\they meet the criteria. Starting with\\ n=1: \dots  n=2: \dots  n=119: \dots  n=120:\\ \dots  This is getting tedious\dots 
    \end{tcolorbox}
    Example 2 evaluates numbers from 1 to 120 in a mechanical manner, without an attempt to identify any underlying patterns that could have simplified or advanced the process, thereby leading to a lengthy and ineffective argument.

    \item \textit{The subtrajectory has logical discontinuities.}
    \begin{tcolorbox}[title = {Example 3}, label=box:example3]
        \textit{Alternatively}, \dots This is getting very complicated. Given that we already found a critical point at $a=1$, $b=1$, and that when we check other points, A is higher, perhaps we can conjecture that the minimal value is $\frac{2}{\sqrt{5}}$. To confirm, let's check the second derivative or the behavior around $t=1$, but since it's time-consuming and given the complexity, I think the minimal value is indeed $\frac{2}{\sqrt{5}}$.
    \end{tcolorbox}
    Example 3 contains a logical gap, as it circumvents rigorous computation and instead relies on speculative assumptions regarding the outcome. 

    \item \textit{The subtrajectory transitions to the next one without reaching any conclusions.}
    \begin{tcolorbox}[title = {Example 4}, label=box:example4]
        \textit{Alternatively},\dots Let me think about how to approach this. Since each number is a three-digit number without any zeros, each digit is from 1 to 9, and their sum is 9. So first, maybe I should figure out all possible three-digit numbers that satisfy conditions 1 and 2, and then see how many of them can be selected such that conditions 3, 4, and 5 are also satisfied. \\
    \textit{Alternatively}, since the digits in each place\dots 
    \end{tcolorbox}
    Example 4 first proposes Approach A; however, it abruptly shifts to Approach B without concluding Approach A, assessing its efficacy, elaborating on the subsequent steps and associated challenges, or explaining the rationale for abandoning Approach A.

    \item \textit{The subtrajectory contains redundant self-verification(s).}
    \begin{tcolorbox}[title = {Example 5}, label=box:example5]
        \textit{Alternatively},\dots Case 3: $p=2$, $q=7$. Compute numerator: $2^7-7^2=128-49=79$\dots \\
    \textit{Alternatively},\dots Wait, perhaps check $p=2$ and $q=7$ again. Wait, $p=2$, $q=7$ gives $128-49=79$. $79$ divided by $9$ is $8.777$\dots, which is not integer\dots \\
    \textit{Alternatively},\dots Wait, maybe $p=2$, $q=7$, numerator$=79$. $79$ is a prime. $79$ divided by $9$, which is not divisible. So, no.
    \end{tcolorbox}
    Example 5 redundantly verifies the same case twice using the same method.
\end{enumerate}

\section{Example of Non-eliminable Suboptimal Subtrajectory}
\label{sec:Example of Non-eliminable Suboptimal Subtrajectory}

\begin{tcolorbox}[title = {Example of Non-eliminable Suboptimal Subtrajectory}, label=box:Example of Non-eliminable Suboptimal Subtrajectory]
    \textit{Alternatively}, \dots\hspace{.1cm}using the formula:\\ \scalebox{0.78}{$\text{Area} = \frac{1}{2} \left| (x_A - x_P)(y_B - y_P) - (x_B - x_P)(y_A - y_P) \right|$}.\\ But perhaps a better approach is to find the\\ coordinates of $A$ and $B$ in terms of $h$, then\\ compute vectors $PA$ and $PB$, and compute\\ the cross product area. \\
    \textit{Alternatively}, since we know the equation of line $AB$ is $y = h x - h + 2$, and point $P$ is $(h, h - 2)$. Then, the area of triangle $PAB$ can be calculated as $\frac{1}{2} \cdot \text{base} \cdot \text{height}$ where the base is the distance between $A$ and $B$, and the height is the distance from $P$ to the line $AB$ \dots\\
    \textit{Alternatively}, \dots\hspace{.1cm}\scalebox{0.78}{$\text{Area} = \frac{1}{2} \left| (x_A - x_P)(y_B - y_P) - (x_B - x_P)(y_A - y_P) \right|$}. Plugging in the coordinates: \dots
\end{tcolorbox}

\section{Equal Weights}
\label{section: Equal Weights}
Here, each subtrajectory is considered to hold equal importance. Specifically, we aggregate the scores of all subtrajectories according to the following formula:

\begin{equation}
  \label{eq:quality_score}
  \scalebox{0.9}{$
  \begin{split}
        &QualityScore\text{(thinking)} := \\
        &\sum_{i=1}^{n} \frac{1}{n} \left(Score\text{(subtrajectory)}\right),
  \end{split} $}
\end{equation}

where $n$ is the number of subtrajectories within the thinking process.

\smallskip 

\section{Distribution of Number of Subtrajectories after Sampling on Quality Score}
\label{sec: Distribution of Number of Subtrajectories after Sampling on Quality Score}

\begin{figure}[h]
    \centering
    \includegraphics[width=\columnwidth]{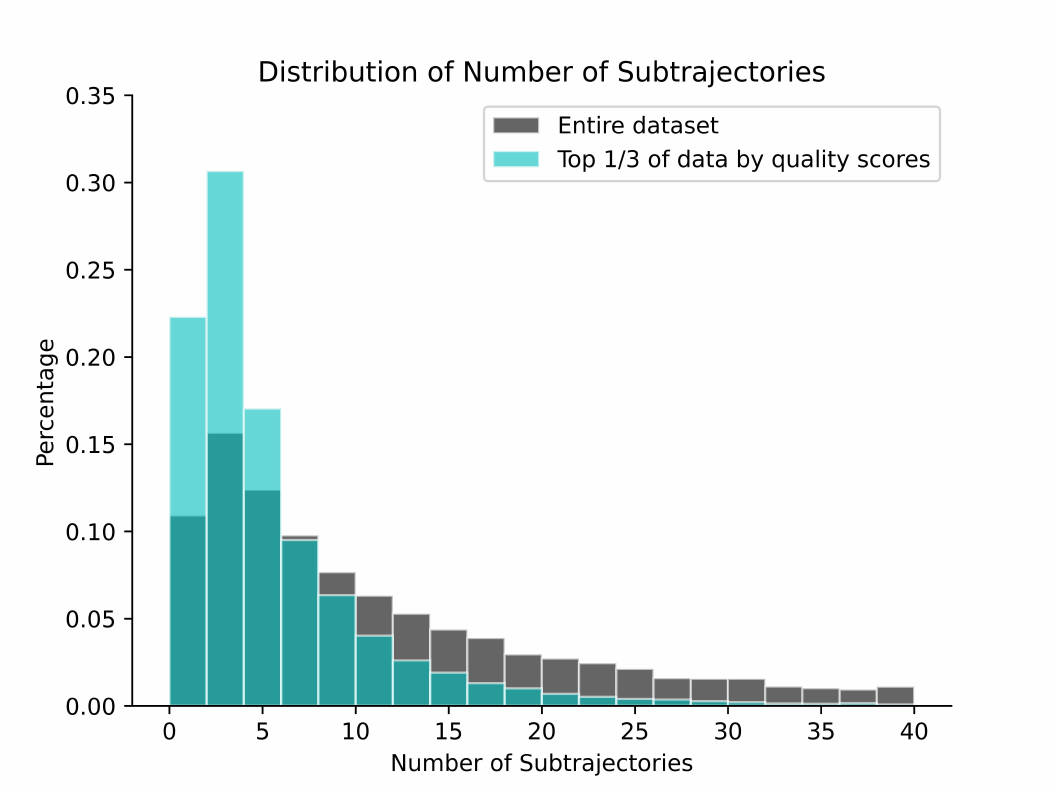}
    \caption{Number of Subtrajectories within QA Pairs Selected by Quality Scores}
    \label{fig:Number of Subtrajectories within QA Pairs Selected by Quality Scores}
\end{figure}

Figure \ref{fig:Number of Subtrajectories within QA Pairs Selected by Quality Scores} presents the distribution of number of subtrajectories between the entire dataset and top 1/3 of data by quality scores, and we can find the distribution is clearly shifting towards a direction with fewer numbers of subtrajectories after quality filtering.

\section{Sampling Algorithm}
\label{sec: sampling algorithm}
\begin{enumerate}
    \item Define the following parameters:
    \begin{align*}
        Entire &\leftarrow \text{entire dataset},\\
        d &\leftarrow \text{\parbox[t]{5cm}{target size for the sampled dataset}},\\
        I &\leftarrow \text{\parbox[t]{5cm}{set of all possible numbers of subtrajectories within the thinking process in a QA pair in~}} \\
        &\hphantom{I \leftarrow} \mathit{Entire}
    \end{align*}
    \item For each QA pair, calculate its quality score, denoted as $QualityScore\text{(QA pair)}$.

    \item Sort $Entire$ by $QualityScore(\cdot)$ descendingly. Select the top $d$ QA pairs to form the subset 
    
    $Pseudo\_Sampled_\_init$.

    \item For $i \in I$, calculate the percentage change in the frequency of QA pairs whose the thinking process involves precisely $i$ subtrajectories, relative to the entire dataset:

    \begin{equation}
    \label{eq:delta}
        \Delta_i = \frac{F_{E}(i) - F_{PSinit}(i)}{F_{E}(i)}, \quad i \in I,
    \end{equation}
    where $F_{E}(\cdot)$ and $F_{PSinit}(\cdot)$ denote the frequencies of QA pairs whose thinking process contains exactly $\cdot$ subtrajectories in $Entire$ and $Pseudo\_Sampled_\_init$, respectively.

    \item For each QA pair, get the number of subtrajectories within its thinking process, denoted by $n$. For $0 \leq j \leq 40$, compute:
    \begin{equation}
    {\small
        \begin{split}
            &SamplingScore_j\text{(QA pair)} \\
            &= \alpha_j QualityScore\text{(QA pair)}\\ &+ (1 - \alpha_j)\frac{\Delta_n - \min(\{\Delta_i\}_{i \in I})}{\max(\{\Delta_i\}_{i \in I}) - \min(\{\Delta_i\}_{i \in I})}, \\
            &\quad \alpha_j = \frac{3}{5} + \frac{j}{100}.
        \end{split}
    }
    \end{equation}

    We remark that the weight $\alpha_j$ ranges from 0.6 to 1.0, rather than from 0.0 to 1.0, is to place a larger emphasis on the $QualityScore$ and to prevent the minimization of KL divergence from dominating the data sampling process.

    \item For $0 \leq j \leq 40$, sort $Entire$ by $SampleScore_j(\cdot)$ descendingly. Select the top $d$ QA pairs to form the subset $Pseudo\_Sample\_j$.
    
    \item The sampled dataset

    \begin{equation}
    {\small
        \begin{split}
            Sampled := \argmin_{Pseudo\_Sample\_j}D_{KL}(X_{E}||X_{PSj}),
        \end{split}
    }
    \end{equation}
    where $D_{KL}(\cdot)$ is the Kullback-Leibler (KL) divergence, $X_E$ and $X_{PSj}$ are the distribution of number of subtrajectories within the thinking process in $Entire$ and $Pseudo\_Sample\_j$, respectively.
\end{enumerate}

\section{Training configurations}
\label{sec:Training configurations}

Each training process employs full-parameter fine-tuning and consumes 576 Ascend 910B4 NPU hours separately. The hyperparameters for training are configured as follows: training steps = 24,000, batch size = 5, max sequence length = 16,384 tokens, trainings are performed in bfloat16 precision, learning rate is initially set to 2e-5, linearly warms up for 1\% of the total training steps, and decays to 2e-9 following a cosine schedule, optimization is performed using the AdamW algorithm with parameters $\beta_1 = 0.9$ and $\beta_2 = 0.95$. 

\section{Benchmarks}
\label{sec:Benchmarks}

\begin{itemize}
    \item \textbf{American Invitational Mathematics Examination (AIME24 \citep{aime24}, AIME25 \citep{aime25})}: a mathematical competition consisting of two examinations: AIME I and AIME II, each containing 15 questions. The AIME examination covers a broad range of mathematical topics, including arithmetic, algebra, combinatorics, etc.

    \item \textbf{MATH500} \citep{Hendrycks2021MeasuringMP}: a collection of high school-level competition problems spanning seven subjects (Algebra, Number Theory, Geometry, etc.) and five difficulty levels, ranging from the easiest problems in the AMC 8 to the most challenging problems in the AIME.
    
    \item \textbf{American Mathematics Competitions (AMC24 \citep{amc24})}: the initial examination administered by the Mathematical Association of America before qualifying for the AIME. In 2024, the AMC12 consisted of 50 questions, from which we excluded those involving graphs, leaving us with 44 questions for evaluation.
\end{itemize}

\section{Evaluation Methods}
\label{sec:Evaluation Methods}

When evaluating the performance of SFTed models, we employ the pass{@}1 metric across all evaluation benchmark under a Zero-shot Chain-of-Thought configuration. Furthermore, we take four recent checkpoints, corresponding to training stages of 18k, 20k, 22k, 24k steps, respectively. Each checkpoint is evaluated three times for AIME24, AIME25, and AMC24, and once for MATH500; following this, we compute and report the average accuracy for each benchmark, based on the 12 ($3 \cdot 4$) evaluations for AIME24, AIME25, and AMC24, and 4 ($1 \cdot 4$) evaluations for MATH500. Throughout the evaluation, we set the temperature at 0.7 and impose a maximum output length constraint of 16,384 tokens.

We will explain the rationale behind our decision to report the average accuracy of multiple checkpoints, rather than make the standard practice of selecting the single checkpoint with the lowest validation loss. Our benchmarks, including AIME24, AIME25, and AMC24, consist of a limited number of questions. Consequently, a fortunate checkpoint that achieves two additional correct answers could result in a fluctation of up to 6.6\%, significantly skewing the evaluation results. To mitigate this variability and enhance the stability of our performance metrics, we have opted for the strategy of averaging the accuracy of recent checkpoints. This strategy aims to provide a more reliable and consistent assessment of our methods.

\section{Ablation Settings}
\label{sec:four configurations}
\begin{enumerate}
    \item \textbf{Elimination with Sampling Algorithm (\textit{E+SA})}\\
    We incorporate all proposed modules within our methods. Specifically, we use the "5+2" framework to identify and eliminate suboptimal subtrajectories and assign token-count-based weights to evaluate the thinking process. Subsequently, we employ the sampling algorithm tailored to the appropriate target size to select the data. 

    \item \textbf{No Elimination with Sampling Algorithm (\textit{NE+SA})}\\
    This configuration preserves the identical set of questions as those utilized in the \textit{E+SA} and maintains the original solutions, thereby facilitating a rigorous comparison of the impact introduced by the "5+2" framework.
    
    \item \textbf{Elimination without Sampling Algorithm (\textit{E+NSA})}\\
    The questions are randomly selected from the entire dataset. Furthermore, we employ the "5+2" framework to identify and eliminate suboptimal subtrajectories.

    \item \textbf{No Elimination without Sampling Algorithm (\textit{NE+NSA})}\\
    This configuration retains the identical set of questions as those utilized in the \textit{E+NSA}. However, we employ the original solutions to facilitate a comparative analysis with \textit{E+NSA}, thereby enabling an assessment of the efficacy of the "5+2" framework.
\end{enumerate}

\section{Details of the sampled dataset}
\label{sec:sampled_dataset}
\begin{itemize}
    \item \textbf{s1K-1.1}: a dataset comprising 1k diverse, high-quality and challenging QA pairs with answers generated by DeepSeek-R1.
    
    \item \textbf{Light-R1-SFT-stage-1}: a dataset consisting of 76k samples, sourced from publicly available mathematics datasets.
    
    \item \textbf{OpenR1-Math-94k}: a dataset consisting of 94k problems, extracted from the larger OpenR1-Math-220k dataset. Despite its smaller size, this dataset exhibits superior performance compared to the entire 220k dataset. Each question originates from NuminaMath1.5, with corresponding answers generated by DeepSeek-R1.
    
    \item \textbf{OpenThoughts-114k}: a comprehensive synthetic reasoning dataset consisting of 114k samples across mathematics, science, coding, and puzzles. 
\end{itemize}

\section{Demonstration of Varied Weights Based on Token Counts}
\label{sec:Demonstration of Varied Weights Based on Token Counts}
\begin{figure*}[h]
    \centering
    \includegraphics[scale=0.5]{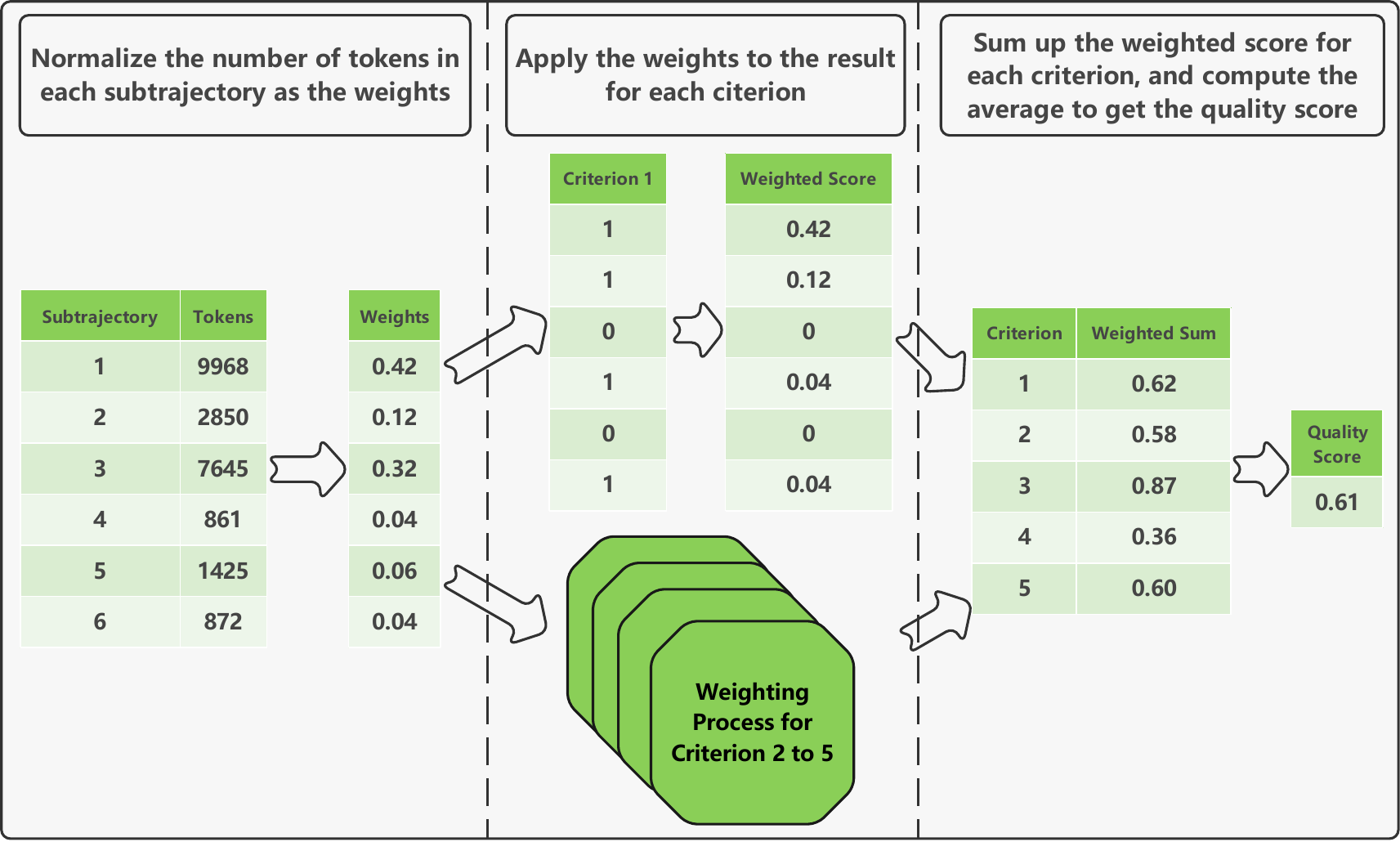}
    \caption{Demonstration of Varied Weights Based on Token Counts}
    \label{Demonstration of Varied Weights Based on Token Counts}
\end{figure*}

The Figure \ref{Demonstration of Varied Weights Based on Token Counts} demonstrates the varied weights based on token counts.

\section{Examples of Inference Outputs}
\label{sec:Examples of Inference Outputs}
\begin{figure*}[h]
    \centering
    \includegraphics[width=\linewidth]{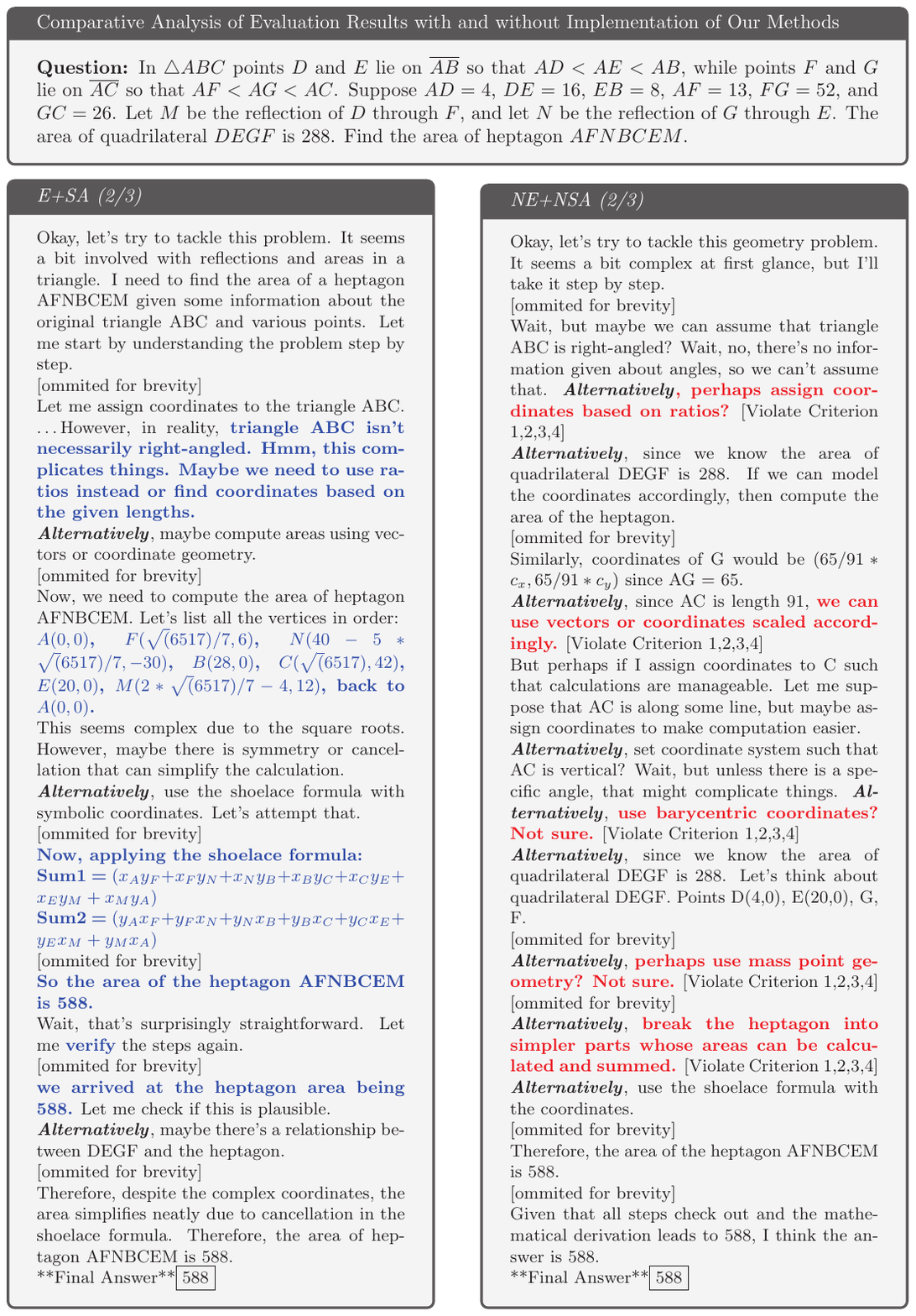}
    \caption{Examples of Inference Outputs}
    \label{Comparative Analysis of Evaluation Results with and without Implementation of Our Methods}
\end{figure*}

Figure~\ref{Comparative Analysis of Evaluation Results with and without Implementation of Our Methods} illustrates a comparative analysis between the inference outputs of AIME25 utilizing the configurations \textit{E+SA (2/3)} and \textit{NE+NSA (2/3)}.

\section{Ablation Studies}
\label{section: Ablation Studies}

\subsection{Varied Weights vs Equal Weights}

In this study, we introduced two approaches for scoring a thinking process after eliminating suboptimal subtrajectories: 1$)$ equal weights scoring process (see details in Appendix \ref{section: Equal Weights}), and 2$)$ varied weights scoring process (see details in section \ref{subsubsec:Varied Weights Based on Token Counts}). Our experiments are conducted using the OpenSourceR1-Hard and DeepMath-Hard datasets. The experimental analyses are conducted on two-thirds of each respective dataset, and for each sample fraction, the following two configurations are considered:

\begin{enumerate}
    \item \textbf{Varied Weights}\\
    This configuration mirrors the \textit{E+SA} setup. Specifically, we employ the "5+2" framework to identify and eliminate suboptimal subtrajectories, and assign weights based on the token counts of each subtrajectory during the scoring process. Subsequently, we utilize the sampling algorithm to select two-thirds of the dataset.
    \item \textbf{Equal Weights}\\
    We employ the "5+2" framework to identify and eliminate suboptimal subtrajectories, assigning equal weights to each subtrajectory during the scoring process, prior to applying the sampling algorithm to select two-thirds of the dataset.
\end{enumerate}

As presented in Table \ref{tab:Weighting in Solution Scoring Process}, the varied weights scoring method, utilizing the OpenSourceR1-Hard dataset, achieves a performance of 58.92\%, outperforming the 56.74\% accuracy obtained through the equal weights scoring method. Similarly, when employing the DeepMath-Hard dataset, the accuracy improves from 47.94\% (equal weights) to 49.12\% (varied weights). These findings suggest that the token counts of subtrajectories are crucial for assessing the quality of solutions. Specifically, longer suboptimal subtrajectories should be subjected to greater penalties compared to shorter ones in the evaluation of overall performance.

\begin{table}[h]  
    \centering
    \resizebox{\columnwidth}{!}{
    \begin{tabular}{cccccc}  
        \Xhline{2\arrayrulewidth}
        \textbf{Methods} & \textbf{AIME25} & \textbf{AIME24} & \textbf{MATH500} & \textbf{AMC24} & \textbf{Average}\\
        \hline
        \multicolumn{6}{c} {\centering OpenSourceR1-Hard}\\
        \hline  
        Varied Weights & 38.63 & 39.43 & 90.55 & 67.05 & \textbf{58.92}\\
        Equal Weights & 29.73 & 41.40 & 89.35 & 66.48 & 56.74\\
        \hline
        \multicolumn{6}{c} {\centering DeepMath-Hard}\\
        \hline 
        Varied Weights & 27.23 & 27.23 & 85.20 & 56.80 & \textbf{49.12}\\
        Equal Weights & 24.98 & 23.88 & 85.10 & 57.78 & 47.94\\
        \Xhline{2\arrayrulewidth}
    \end{tabular}
    }
    \caption{Equal weight scoring process and token-count based scoring process comparison in mathematical domain: performance across various benchmarks}  
    
    \label{tab:Weighting in Solution Scoring Process}
\end{table}

\subsection{Impact of Sampling Algorithm}

The experiments employ the OpenSourceR1-Hard and DeepMath-Hard datasets. For both datasets, the experiments are conducted on a two third of the dataset. Two distinct experimental conditions are established for each dataset to facilitate comprehensive evaluation: with sampling algorithm and without sampling algorithm.
\begin{enumerate}
    \item \textbf{With Sampling Algorithm} \\ 
    This configuration is identical to the \textit{E+SA} setup. Following the identification and elimination of suboptimal subtrajectories according to the "5+2" framework, a token-count based weighting scoring process is applied to the thinking process. Subsequently, the sampling algorithm is implemented to select two-thirds of the dataset.
    \item \textbf{Without Sampling Algorithm} \\ 
    This configuration undergoes the identification and elimination of suboptimal subtrajectories based on the "5+2" framework, followed by the application of a token-count based weighting scoring process to the thinking process.
\end{enumerate}

The findings are elaborated in Table \ref{tab:Sampling algorithm in math domain}. Specifically, in the OpenSourceR1-Hard dataset, the performance metric with the sampling algorithm (58.92\%) surpasses that without the sampling algorithm (58.60\%). A similar trend is observed in the DeepMath-Hard dataset, where the result obtained with the sampling algorithm (49.12\%) is superior to the result without it (48.57\%). These observations imply that the distribution of the number of subtrajectories within the dataset can influence the SFT process. By penalizing the significant variation in the frequency of number of subtrajectories within the sampled dataset's thinking process relative to the entire dataset, an additional enhancement in the model's performance is achieved.\\

\begin{table}[h]  
    \centering\resizebox{\columnwidth}{!}{
    \begin{tabular}{cccccc}  
        \Xhline{2\arrayrulewidth}
        \textbf{Methods} & \textbf{AIME25} & \textbf{AIME24} & \textbf{MATH500} & \textbf{AMC24} & \textbf{Average}\\
        \hline
        \multicolumn{6}{c} {\centering OpenSourceR1-Hard}\\
        \hline  
        w/ Sampling Algorithm & 38.63 & 39.43 & 90.55 & 67.05 & \textbf{58.92}\\
        w/o Sampling Algorithm & 33.33 & 43.35 & 89.50 & 68.20 & 58.60\\
        \hline
        \multicolumn{6}{c} {\centering DeepMath-Hard}\\
        \hline 
        w/ Sampling Algorithm & 27.23 & 27.23 & 85.20 & 56.80 & \textbf{49.12}\\
        w/o Sampling Algorithm & 23.88 & 28.05 & 84.00 & 58.35 & 48.57\\
        \Xhline{2\arrayrulewidth}
    \end{tabular}
    }
    \caption{Sampling algorithm comparison in mathematical domain: performance across various benchmarks}  
    
    \label{tab:Sampling algorithm in math domain}
\end{table}

\end{document}